\documentclass[conference]{IEEEtran}
\IEEEoverridecommandlockouts
\usepackage{cite}
\usepackage{amsmath,amssymb,amsfonts}
\usepackage{algorithmic}
\usepackage{graphicx}
\usepackage{textcomp}
\usepackage{xcolor}
\usepackage{url}
\usepackage{graphicx}
\usepackage{array}
\usepackage{balance}
\usepackage{booktabs} 
\usepackage{multirow}

\makeatletter
\newcommand{\linebreakand}{%
  \end{@IEEEauthorhalign}
  \hfill\mbox{}\par
  \mbox{}\hfill\begin{@IEEEauthorhalign}
}
\makeatother

\def\BibTeX{{\rm B\kern-.05em{\sc i\kern-.025em b}\kern-.08em
    T\kern-.1667em\lower.7ex\hbox{E}\kern-.125emX}}
    
\newcolumntype{G}{>{\centering\arraybackslash}m{5in}}

\begin{document}

\title{A Pre-trained Conditional Transformer for Target-specific De Novo Molecular Generation}

\author{\hspace{-20pt}\IEEEauthorblockN{Wenlu Wang\textsuperscript{\textsection}}
\IEEEauthorblockA{
\textit{\hspace{-20pt}Texas A\&M University-Corpus Christi}\\
\hspace{-20pt}Corpus Christi, TX \\
\hspace{-20pt}wenlu.wang@tamucc.edu}
\and
\IEEEauthorblockN{\hspace{-30pt}Ye Wang\textsuperscript{\textsection}}
\IEEEauthorblockA{
\hspace{-30pt}\textit{Biogen}\\
\hspace{-30pt}Cambridge, MA \\
\hspace{-30pt}ye.wang@biogen.com}
\linebreakand%
\IEEEauthorblockN{\hspace{30pt}Honggang Zhao}
\IEEEauthorblockA{
\hspace{30pt}\textit{Cornell University}\\
\hspace{30pt}Ithaca, NY \\
\hspace{30pt}hz269@cornell.edu}
\and
\IEEEauthorblockN{Simone Sciabola}
\IEEEauthorblockA{
\textit{Biogen}\\
Cambridge, MA \\
simone.sciabola@biogen.com}
}

\maketitle
\begingroup\renewcommand\thefootnote{\textsection}
\footnotetext{Equal contribution}
\endgroup

\begin{abstract}

In the scope of drug discovery, the molecular design aims to identify novel compounds from the chemical space where the potential drug-like molecules are estimated to be in the order of $10^{60}–10^{100}$. 
Since this search task is computationally intractable due to the unbounded search space, deep learning draws a lot of attention as a new way of generating unseen molecules. As we seek compounds with specific target proteins, we propose a Transformer-based deep model for de novo target-specific molecular design. The proposed method is capable of generating both drug-like compounds (without specified targets) and target‐specific compounds. The latter are generated by enforcing different keys and values of the multi-head attention for each target. In this way, we allow the generation of SMILES strings to be conditional on the specified target. Experimental results demonstrate that our method is capable of generating both valid drug-like compounds and target-specific compounds. Moreover, the sampled compounds from conditional model largely occupy the real target-specific molecules' chemical space and also cover a significant fraction of novel compounds. 

\end{abstract}


\section{Introduction}

Small molecule drug design aims to identify novel compounds with desired chemical properties. From the computational perspective, we consider this task an optimization problem, where we search for the compounds that will maximize our quantitative goals in chemical space. However, this optimization task is computationally intractable because of the unbounded search space. It has been estimated that the range of potential drug-like molecules estimated to be in the order of $10^{60}$ to $10^{100}$\cite{schneider2005computer}, but only about $10^8$ molecules have ever been synthesized\cite{kim2016pubchem}. It is impossible to test all the possible virtual molecules due to the combinatorial explosion: the number of different ways the elements are linked is enormous. Numerous computational methods, such as virtual screening, combinatorial libraries, and evolutionary algorithms, have been developed to search the chemical space in silico and in vitro. Recent works have demonstrated that machine learning, especially deep learning methods, could produce new small molecules\cite{gomez2018automatic,jin2018junction,zhavoronkov2019deep} with biological activity. We aim to incorporate as much chemical domain knowledge as possible for directed navigation toward a desired location in the search space. 

\begin{figure}[!t]
    \centering
    \vspace{5pt}
    \resizebox{!}{.13\textwidth}{
    \includegraphics{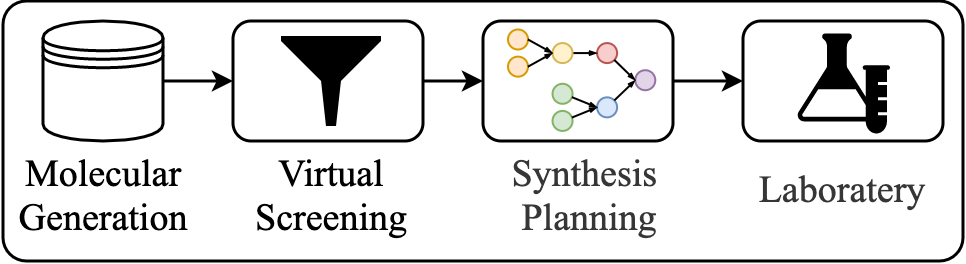}}
    \caption{Deep generative model-based drug discovery workflow.}
    \label{fig:drug-workflow}
\end{figure}

As shown in Figure~\ref{fig:drug-workflow}, deep generative models can generate millions of novel compounds, which can be further refined with the help of virtual screening tools using statistical and physics-based techniques, and a final set of compounds can be synthesized in the laboratory. This paper works on the first step \textbf{Molecular Generation}.

Since the chemical structures can be represented as SMILES (\underline{S}implified \underline{M}olecular \underline{I}nput \underline{L}ine \underline{E}ntry \underline{S}ystem) strings (will be introduced in Section~\ref{sec:smiles}), the small molecule drug design problem can be transformed to a text generation problem in Natural Language Processing (NLP). A number of deep learning techniques have been successfully applied to text generation. For instance, the GPT (Generative Pre-trained Transformer) series~\cite{GPT, GPT2, GPT3} uses an autoregressive language model to produce human-like text by training from massive unlabeled human-written text. The text generated from the GPT model is of high quality and hard to distinguish from human-written content. Similarly, deep models are likely to learn drug-like structures from massive discovered compounds.

\begin{figure*}
    \centering
    \resizebox{!}{.4\textwidth}{
    \includegraphics{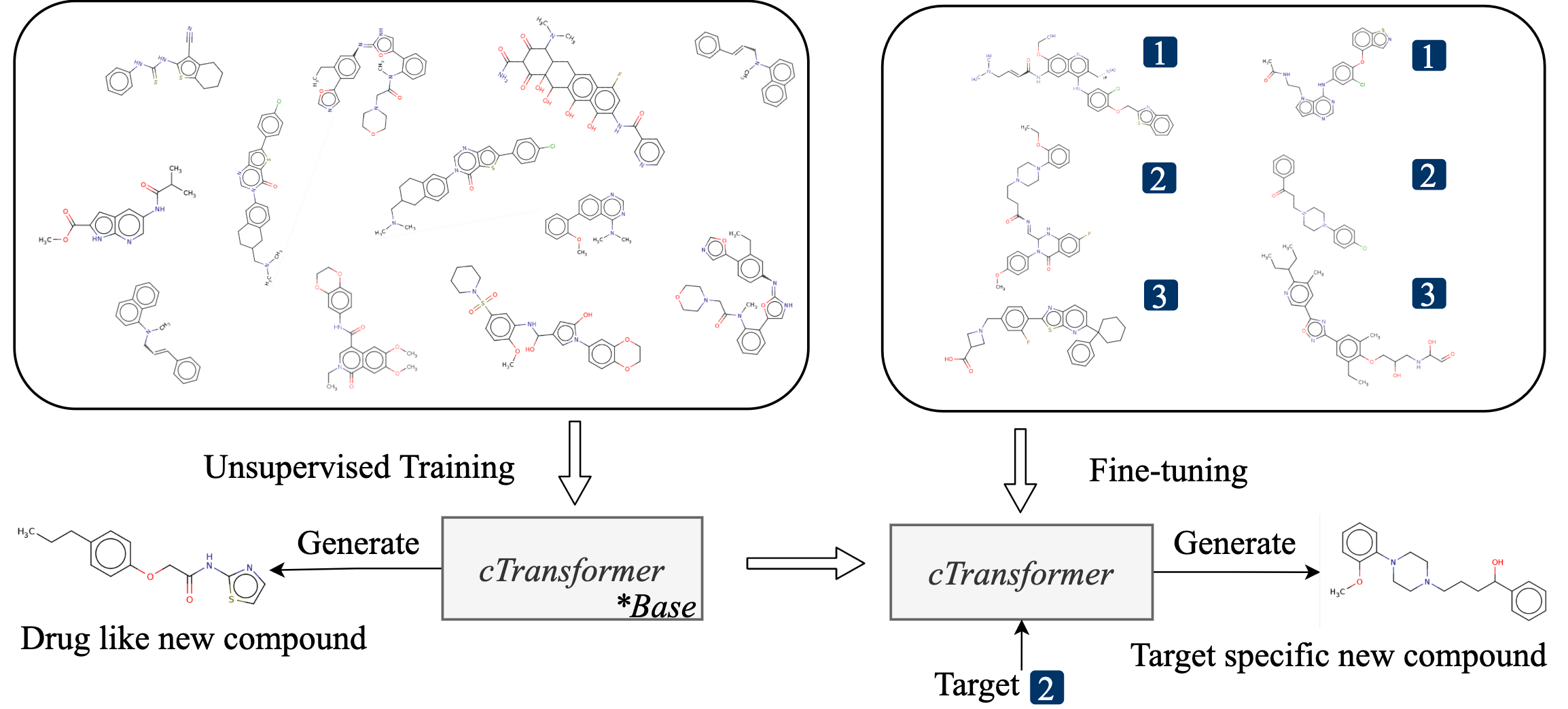}}
    \caption{The workflow of our \textit{cTransformer} design. The base model of \textit{cTransformer} is trained with Table~\ref{tab:data}-(a) data in an unsupervised manner (target-specific embeddings set to \textit{zero}s). The pre-trained \textit{cTransformer} will be further fine-tuned with Table~\ref{tab:data}-(b) data with target-specific embeddings.}
    \label{fig:workflow}
\end{figure*}

Many NLP tasks, such as language translation and text summarization, will be prompted with input (e.g., a sentence and a document), and the problem can be formed as end-to-end learning. Unlike traditional language tasks, drug discovery is a pure generator confined by drug-likeness. The ``drug-like'' structure generally means meeting the a set of criteria, such as good solubility, potency and molecular weight.

In the scope of small molecule drug design, it is essential to enable the generations to be guided by predefined conditions, such as the target protein. Here we formulate the molecular design problem as a conditional sequential generation given the target protein and propose a conditional Transformer architecture \textit{cTransformer} that auto-regressively generates target-specific compounds. As shown in Figure~\ref{fig:workflow}, we first pre-train a transformer on the MOSES~\cite{polykovskiy2020molecular} dataset without target information (denoted as the base model), and after pre-training, the base model can generate drug-like structures. Then, the conditional transformer is fine-tuned on three target-specific datasets (EGFR, HTR1A, and S1PR1 targets). Our experiment results show that the transformer is capable of generating compounds similar to the ones in the training set but are novel structures. We believe the conditional transformer to be a valuable tool for de novo molecule design. Note that our conditional transformer design is not limited to target proteins and can be generalized to a broader scope of targets. 

\begin{table*}[!ht]
    \centering
    \resizebox{!}{.3\textwidth}{
    \begin{tabular}{ >{\centering\arraybackslash}m{2in}  >{\centering\arraybackslash}m{4in}  c}
    \hline
    \hline
    Chemical structure & SMILES &\\
    \hline
\includegraphics[scale = 0.15]{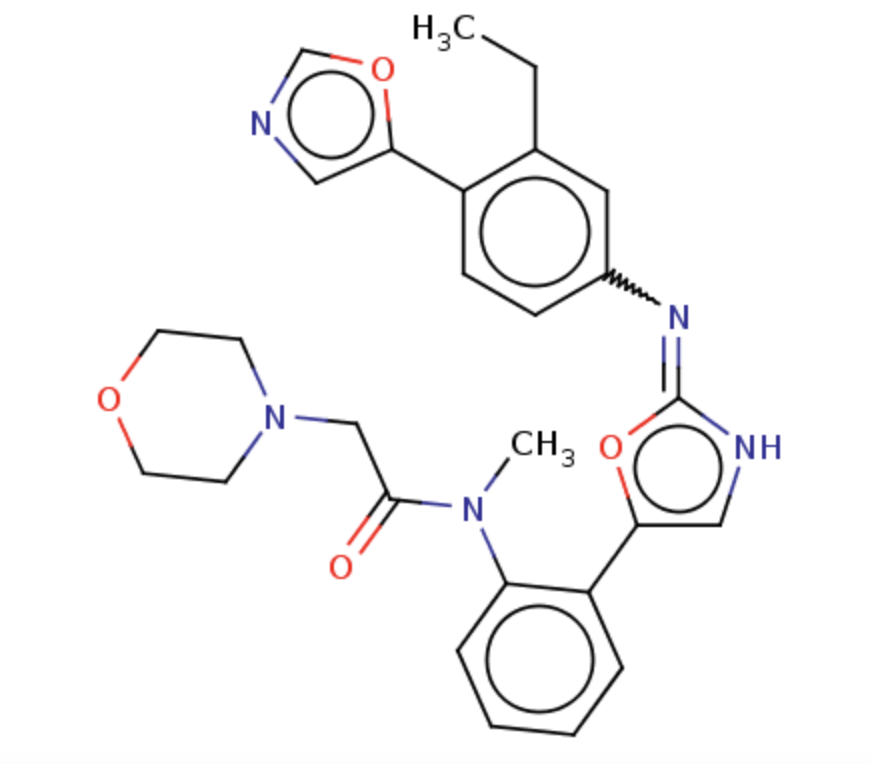} &  CCc1cc(N=c2[nH]cc(-c3ccccc3N(C)C(=O)CN3CCOCC3)o2)ccc1-c1cnco1 & \\[5ex]
\hline
\includegraphics[scale = 0.12]{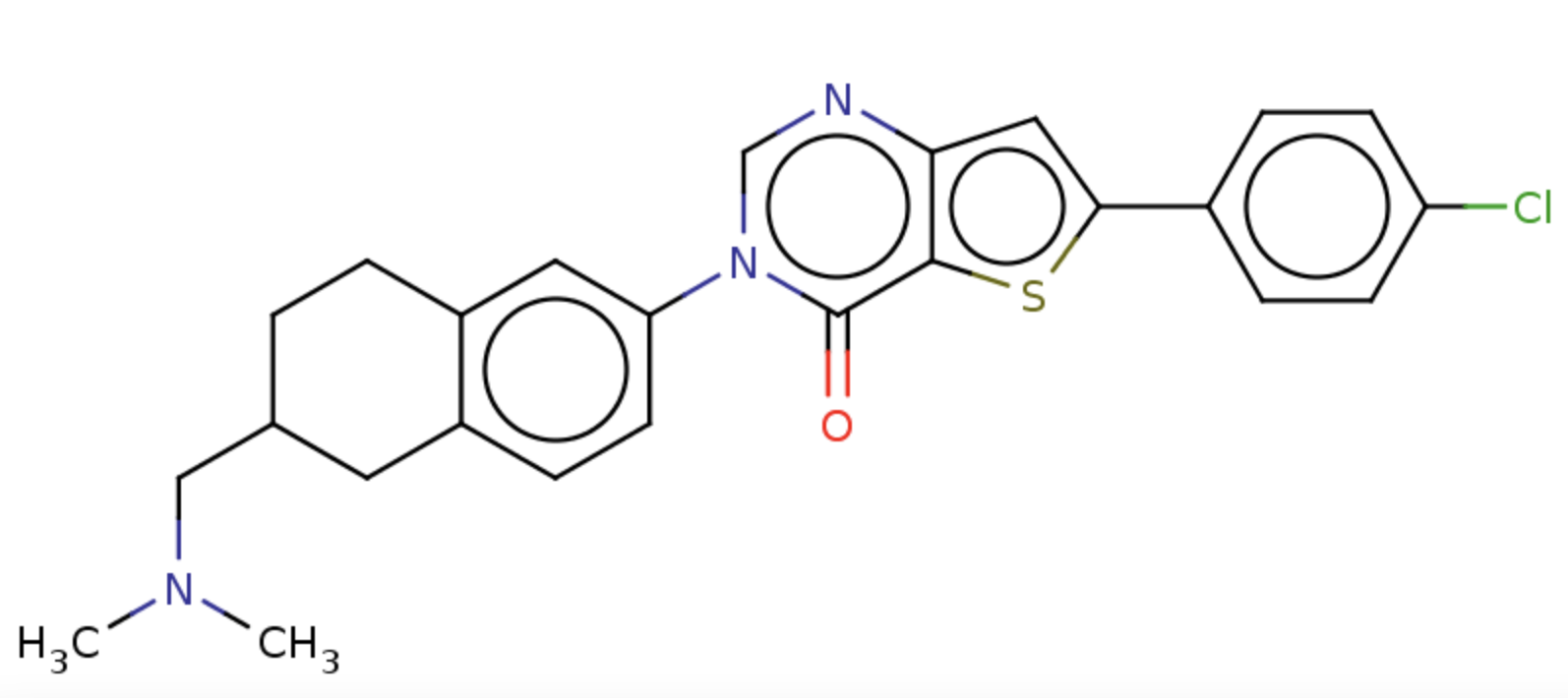} & CN(C)CC1CCc2cc(-n3cnc4cc(-c5ccc(Cl)cc5)sc4c3=O)ccc2C1 & \\
\hline
\includegraphics[scale = 0.12]{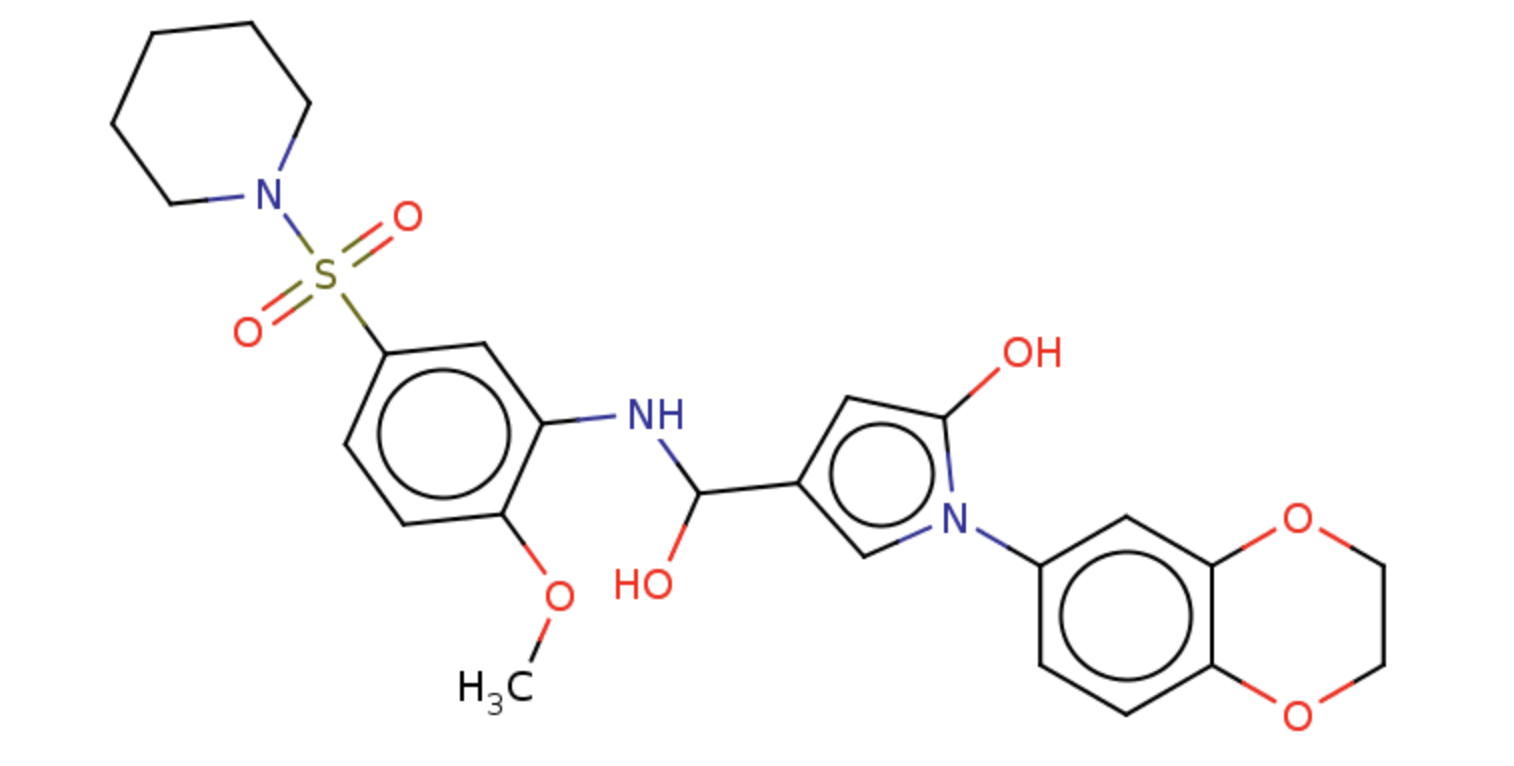} & COc1ccc(S(=O)(=O)N2CCCCC2)cc1NC(O)c1cc(O)n(-c2ccc3c(c2)OCCO3)c1 & \\
\hline
\multicolumn{3}{c}{(a) Pre-training Data.}\\
\\
\\
\hline
\hline 
    Chemical Structure  &  SMILES   &  Target\\
    \hline
    \includegraphics[scale = 0.08]{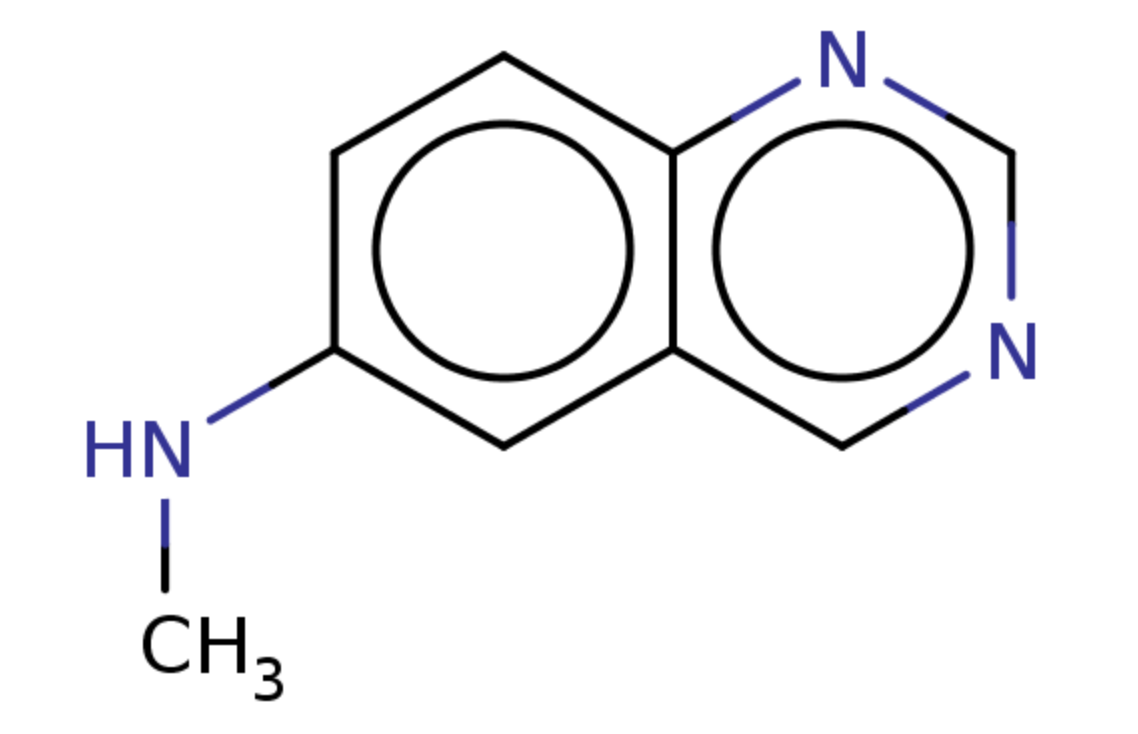} &      CNc1ccc2ncncc2c1 &	EGFR (target 1)\\
    \hline
    \includegraphics[scale = 0.12]{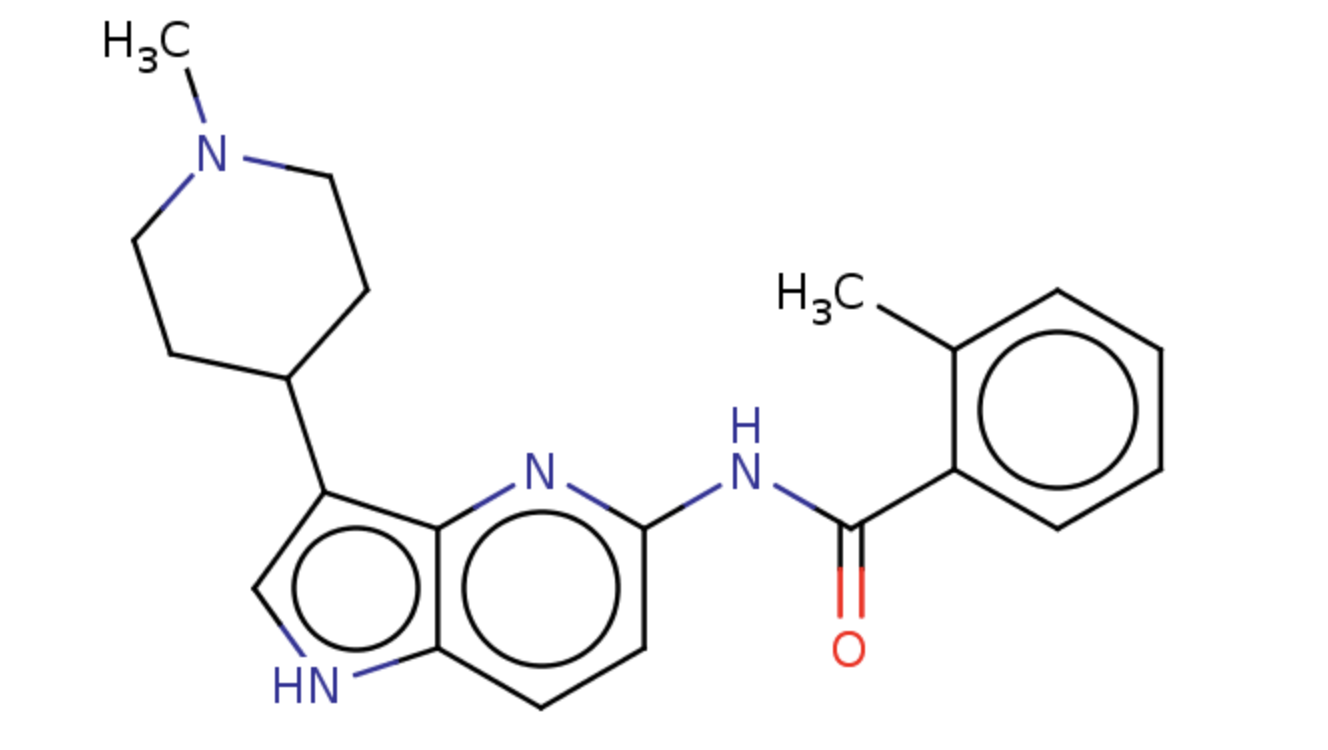} &
      Cc1ccccc1C(=O)Nc1ccc2[nH]cc(C3CCN(C)CC3)c2n1 &	HTR1A (target 2) \\
    \hline
    \includegraphics[scale = 0.13]{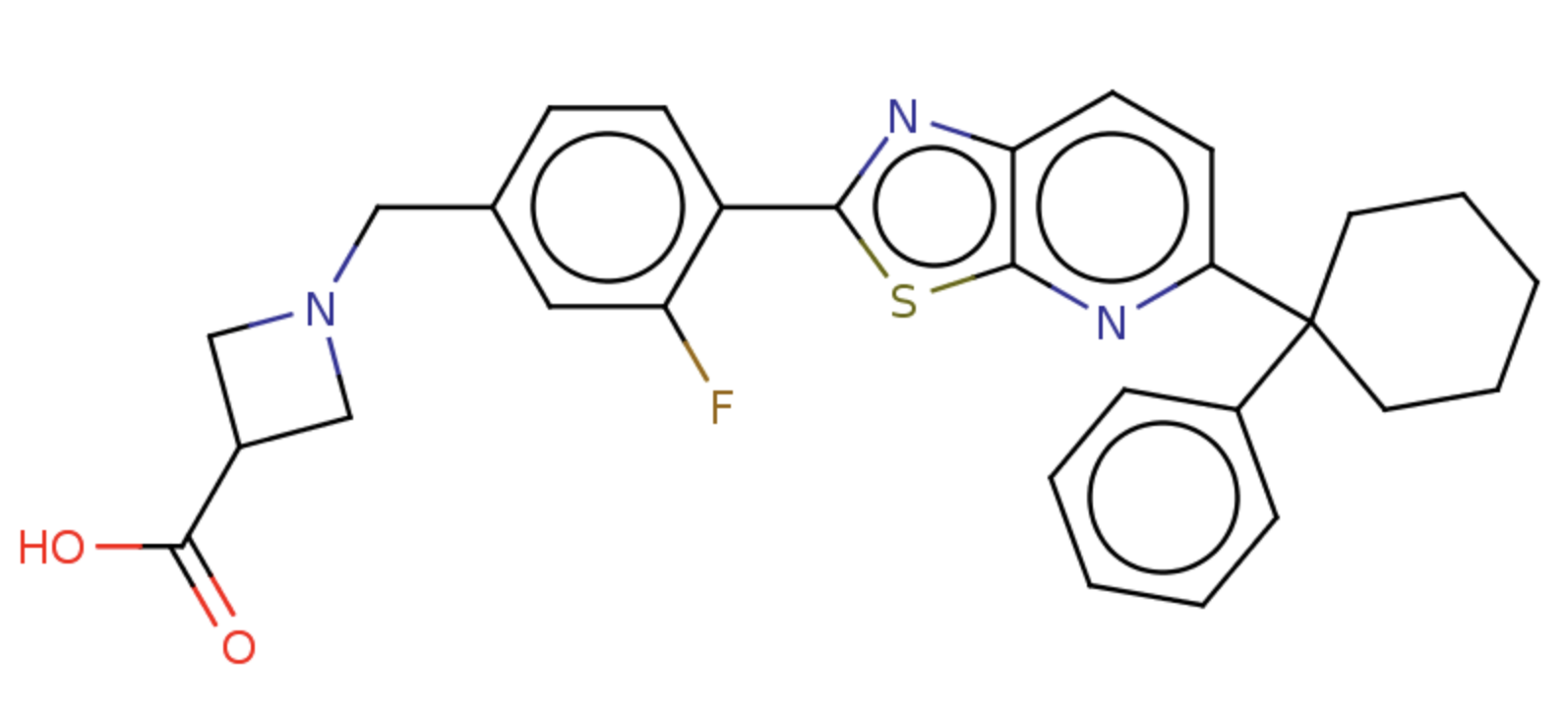} &
      O=C(O)C1CN(Cc2ccc(-c3nc4ccc(C5(c6ccccc6)CCCCC5)nc4s3)c(F)c2)C1 &	S1PR1 (target 3) \\
    \hline
    \multicolumn{3}{c}{(b) Fine-tuning Data.}
    \end{tabular}}
\caption{Data examples.}
\label{tab:data}
\end{table*}

We define our molecular design task as generating a valid SMILES string, which is not covered in training data and has desired potency and physicochemical properties. 
In order to address our task using deep learning, we summarize the challenges as follows:
(1) the generated sequence is a valid drug-like structure in compliance with the SMILES grammar; (2) the generated chemical space covers enough variations that are likely to be new drugs; and (3) the generations can be guided by predefined conditions (incorporating the specified targets).

We propose to use a Transformer-based auto-regressive decoder due to the following reasons: the Transformer structure performs well in sequential modeling, especially for long sequences (challenge (1)), we are able to introduce randomness in the sampling process to generate more variations and makes the trained model more ``creative'' (challenge (2)), and  
we can prompt the Transformer paradigm with different embeddings as \textit{keys} and \textit{values} so that the generative process is conditioned on the specified targets (challenge (3)). The problem setting of molecular generation can be
formed as a text generation problem in natural language processing (NLP) where a generated molecule is a SMILES string, and we enforce conditional generation to enable target-specific generations.

Our contributions are summarized as follows
\begin{itemize}
    \item We propose a conditional generative Transformer \textit{cTransformer} by enforcing different \textit{keys} and \textit{values} of the multi-head attention for each target. We enable the generation of SMILES strings to be conditional on the specified target.
    \item The proposed method is capable of generating both drug-like compounds (without specified targets) and target‐specific compounds. We explore the generated chemical space by sampling and visualizing a large number of novel compounds. The sampled compounds largely occupy the real target-specific data's chemical space and also cover a significant fraction of novel compounds. 
    \item Our conditional generation aims to generate active molecules correspond to EGFR, S1PR1, and HTR1A target protein. We test the target-specific activity of the generated compounds using a computational model (QSAR) for revealing relationships between chemical compounds and biological activities. 
\end{itemize}

\section{Preliminary}\label{sec:pre}

\subsection{SMILES}\label{sec:smiles}

\begin{figure}[!hb]
    \centering
    \resizebox{!}{.17\textwidth}{
    \includegraphics{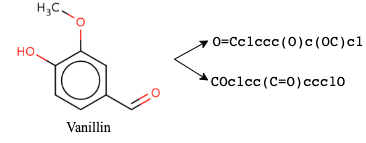}}
    \caption{Ring structures are written by breaking each ring at an arbitrary point to make an acyclic structure and adding numerical ring closure labels to show connectivity between non-adjacent atoms. By choosing different starting points, a molecule might generate different SMILES strings.}
    \label{fig:smiles}
\end{figure}

SMILES (\underline{S}implified \underline{M}olecular \underline{I}nput \underline{L}ine \underline{E}ntry \underline{S}ystem) is a chemical notation that represents a chemical structure by short ASCII strings. One chemical structure might map to different SMILES. There are many softwares that are able to validate a SMILES string and convert SMILES string to two-dimensional drawings.
Figure~\ref{fig:smiles} shows an example of a molecule with its chemical structure and two valid SMILES strings.

\subsection{Data examples}



Drug discovery aims to find novel compounds with specific chemical properties for the treatment of diseases. In addition to generating valid SMILES strings, we need to ensure the corresponding compound is bind to target proteins. We consider the target protein information as an annotated label of each compound. We propose to pre-train on SMILES-only dataset and fine-tune on $<$compound, target protein$>$ pairs with the compound represented as SMILES string.

\noindent (a) Pre-training data: As shown in Table~\ref{tab:data}-(a), we are able to learn the patterns of a valid drug-like SMILES sequence from the pre-training data.

\noindent (b) Fine-tune data: As shown in Table~\ref{tab:data}-(b), the chemical structures are manually annotated with different target proteins. 


\subsection{Problem Definition}

\noindent\textbf{Drug Discovery.} 
The overall idea of drug discovery is to discover new candidate medications~\cite{drugdiscovery}. In the small molecule drug discovery process~\cite{drugdevelopment}, the first step is discovering molecular compounds that potentially have beneficial effects against any diseases. One of the common practices is to screen large-size compound libraries against isolated biological targets~\cite{stokes2020deep}. An intuitive idea is to search for novel molecule candidates \textit{in silico} in the whole drug-like chemical space. Such an idea is not feasible before deep learning due to the unbounded search space of synthetically feasible chemicals. The success of AI changes this line of work since deep generative models are able to learn the distribution of desirable molecules from training data and generate novel drug-like compounds~\cite{luo20213d}.

\vspace{5pt}
\noindent\textbf{AI in Molecular Design.}
Given a set of chemical structures of drugs $\mathcal{D}$, we aim to learn from $\mathcal{D}$ and generate a set of novel chemical structures $\hat{\mathcal{D}}$. $\hat{\mathcal{D}}$ has to be valid drug-like structures and perform well for certain types of metrics in quantitatively measuring the quality of the novel chemical structures. 

\vspace{5pt}
\noindent\textbf{Conditional Molecular Design.}
Appropriate tuning of binding selectivity is a primary objective in the molecular design, and we aim to support conditional generation of novel molecules to be active on the target protein.
In the scope of this paper, we consider the target protein
of the generated molecule and embed the target protein as a condition. Given a condition (e.g., target) $c$, we aim to generate a set of compounds $\hat{\mathcal{D}}_c$ that are more likely to be active on the target protein.

\subsection{Transformer}

RNN-based methods such as seq2seq with attention have achieved great performance in sequential modeling (e.g., machine translation), but the recurring nature of RNN hinders its parallelization, thus making it hard to model long sequences effectively. 
The Transformer~\cite{vaswani2017attention} is proposed to address sequential modeling by attention, which is suitable for parallelization and performs well for long input sequences~\cite{li2019enhancing, jiang2021modeling}. In addition to sequence-to-sequence modeling, Transformer also works well for decoder-only sequence transduction~\cite{DBLP:conf/iclr/LiuSPGSKS18}. 


Many neural sequence transduction models consist of an encoder and a decoder. The encoder first takes a sequence of tokens $(x_1, ..., x_m)$ and transforms them to a sequence of latent representations $\mathbf{z}=(z_1,...,z_m)$ (e.g., memories). The decoder will generate an output sequence $(t_1, ..., t_n)$ one by one conditioning on $\mathbf{z}$. An intuitive way of sequential generation is in an auto-regressive manner~\cite{graves2013generating}, which means consuming all the previously generated tokens while generating the next one. While the traditional Transformer is in an encoder-decoder manner, we define our de novo SMILES generation task as a conditional generator and we use a decoder-only design. Nevertheless, we introduce the complete design of the Transformer to make this paper self-contained.

An attention mechanism mimics the process of querying a set of key-value pairs, where the output is a weighted sum over the values and each weight is based on the matching of the key and query. The multi-head attention projects the keys, values, and queries $h$ times and performs attention in parallel. The formal definition of Multi-head Attention is as follows

\vspace{5pt}
\noindent\textbf{Multi-head Attention (MHA)} We first define some annotations: query matrices $Q_i = QW^Q_i$, key matrices $K_i = KW^K_i$, and value matrices $V_i = VW^V_i$ ($i = 1, ..., h$).

\begin{align*}
    O_i &= Attention(Q_i, K_i, V_i )\\
        &=softmax(\frac{Q_i K^T_i}{\sqrt{d_k}} )V_i\\
\end{align*}

$$MultiHeadAttention(Q, K, V) = \textit{CONCAT}(O_1,…,O_h)W^O$$

$W_i^Q\in \mathbb{R}^{d_{model}\times d_k}, W_i^K \in \mathbb{R}^{d_{model}\times d_k}$
$W_i^V \in \mathbb{R}^{d_{model} \times d_v}$ are learnable parameters.

\vspace{5pt}
\noindent\textbf{Encoder} The encoder has a stack of identical layers. Each has two sub-layers: a multi-head attention component, followed by a feed-forward network; a residual connection is deployed around each of the two sub-layers, followed by layer normalization. 

\vspace{5pt}
\noindent\textbf{Decoder} The decoder also has a stack of identical layers. Each has three sub-layers (two of them are the same as the encoder) including an extra sub-layer performing multi-head attention over the output (e.g., latent representations $\mathbf{z}$) of the encoder. 

\section{Method}




We propose to use a decoder-only Transformer to generate molecules in SMILES format. The token-wise generation is in an auto-regressive manner; at each step, the decoder consumes the previously generated tokens as input while generating the next. 
The proposed model is pre-trained with a large-scale SMILES dataset to learn a parametric probabilistic distribution over the SMILES vocabulary space and ensure the generation is in compliance with the SMILES grammar (e.g., atom type, bond type, and size of molecules) (Section~\ref{sec:pre-train}). Then the conditional generation is enforced by feeding target-specific embeddings to the multi-head attention component of the Transformer (Section~\ref{sec:condition}).


\subsection{Unsupervised Pre-training}~\label{sec:pre-train}

\begin{figure*}
    \centering
    \resizebox{.95\textwidth}{!}{
    \includegraphics{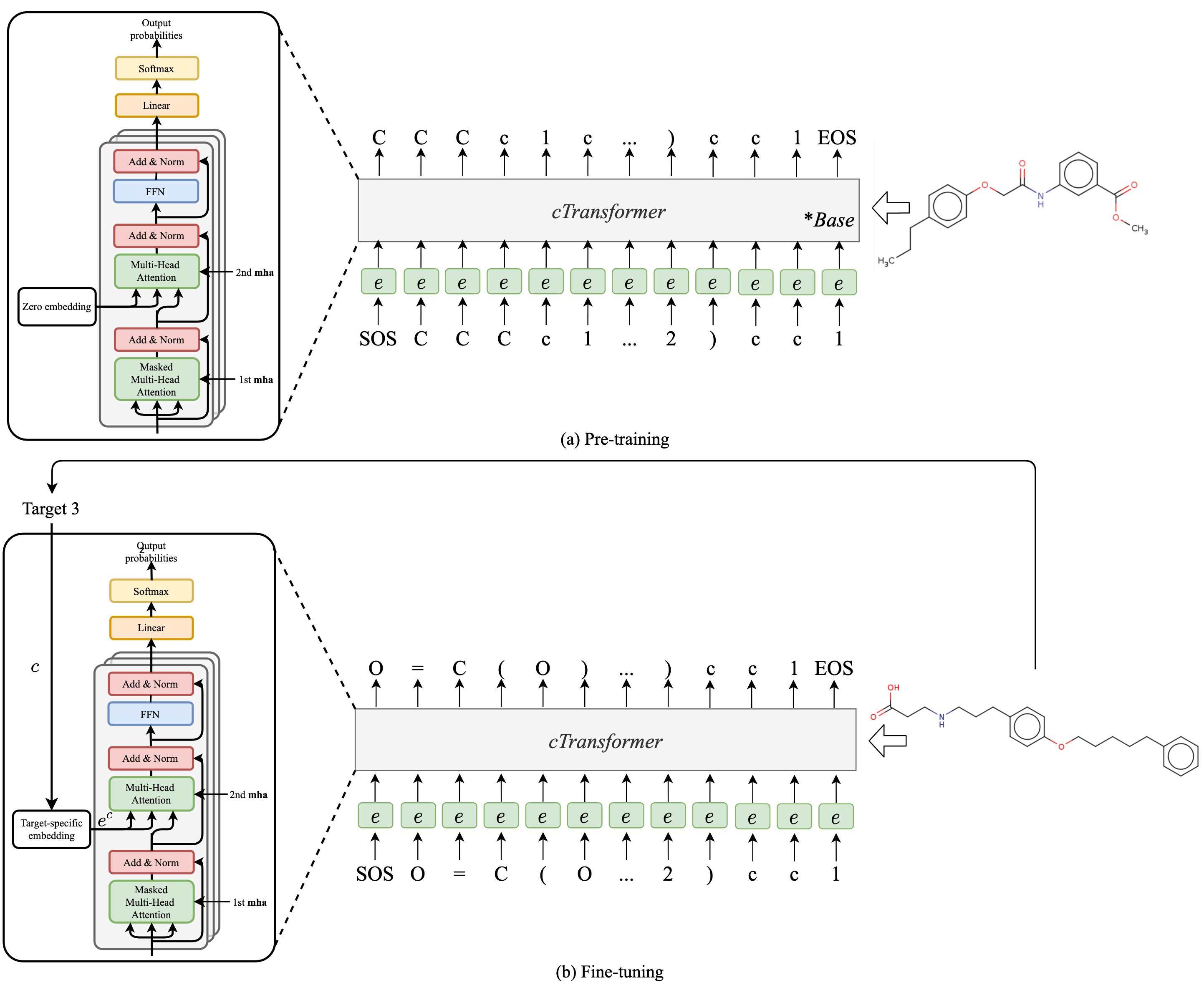}}
    \caption{(a) Pre-training and (b) Fine-tuning.}
    \label{fig:training}
\end{figure*}

Many deep learning tasks rely on supervised learning and human-labeled dataset. For instance, the sequence-to-sequence~\cite{sutskever2014sequence} (or seq2seq) model has been enjoying massive success in many natural language processing applications, and the seq2seq models are usually trained end-to-end with a large number of training pairs (e.g., article-summary pairs for text summarization).

However, in the chemical space, given a large number of unlabeled data but limited labeled data, we form an \textbf{unsupervised learning} task for drug discovery to overcome the challenge of expensive and hard-to-manage human labeling. Instead of a seq2seq encoder-decoder model, we consider a \textbf{decoder-only generative model} and form our task as predicting the next token given previously generated tokens. 

GPT~\cite{GPT} and GPT-2~\cite{GPT2} have gained great success in language generation. Specifically,  GPT-2 learns natural language by predicting next word given the previous words. Inspired by the success of unsupervised Transformer on NLP (e.g., GPT~\cite{GPT} and GPT2~\cite{GPT2}), we propose to use the Transformer-based model for drug discovery. Since a Transformer-based model works well for natural language applications such as writing assistants and dialogue systems, we are optimistic about its capabilities of generating drug-like SMILES sequences. 

We form our drug discovery task by taking advantage of SMILES sequential structure and transforming it as a sequential generation. As shown in Figure~\ref{fig:training}-(a), we formalize our task as predicting the next token given the previous tokens. The sequential dependencies are trained to mimic structures observed in the training set and follow SMILES grammar. During auto-regressive generation, the sampling should be able to produce more variations that are not previously observed. 

\subsection{Conditional Transformer}~\label{sec:condition}

Our \textbf{decoder-only} design is able to memorize drug-like structure from  pre-training. By conditioning on pre-defined conditions, we are able to further confine the search space and sample drug-like structures with desired properties. 


Our decoder learns a parametric probabilistic distribution over the SMILES vocabulary space conditioning on the conditions $c$ (we denote target as $c$) and previously generated tokens. At $i$-th step, our decoder produces a probabilistic distribution of the new token attentively on previously generated token embeddings $\big[e^t_1,...,e^t_{i-1}\big]$ and the target-specific embedding $e^c$.  



\begin{align*}
 z^{(0)} & = \big[e^t_1; e^t_2; ...; e^t_{i-1}\big] \\
 \bar{z}^{(l-1)} & = \textit{LN}(z^{(l-1)} + \textit{MHA}(z^{(l-1)}, z^{(l-1)}, z^{(l-1)}))\\
 \bar{z}^{(l)} & = \textit{LN}(\bar{z}^{(l-1)} + \textit{MHA}(\bar{z}^{(l-1)}, e^c, e^c))\\
 z^{(l)} & = \textit{LN}(\bar{z}^{(l)} + \textit{FFN}(\bar{z}^{(l)}))
\end{align*}

\textit{LN} is layerNorm, \textit{MHA} is multi head attention, and \textit{FFN} is Feed-Forward Networks. Note that we use mask by multiplying masked positions with negative infinity to avoid attending on the masked positions. 
With attending on previous generated tokens, we maintain the structural consistency with the SMILES grammar; In other words, we make sure the generated sequence is drug-like based on the memorization in training.  

\begin{figure*}
    \centering
    \resizebox{.9\textwidth}{!}{
    \includegraphics{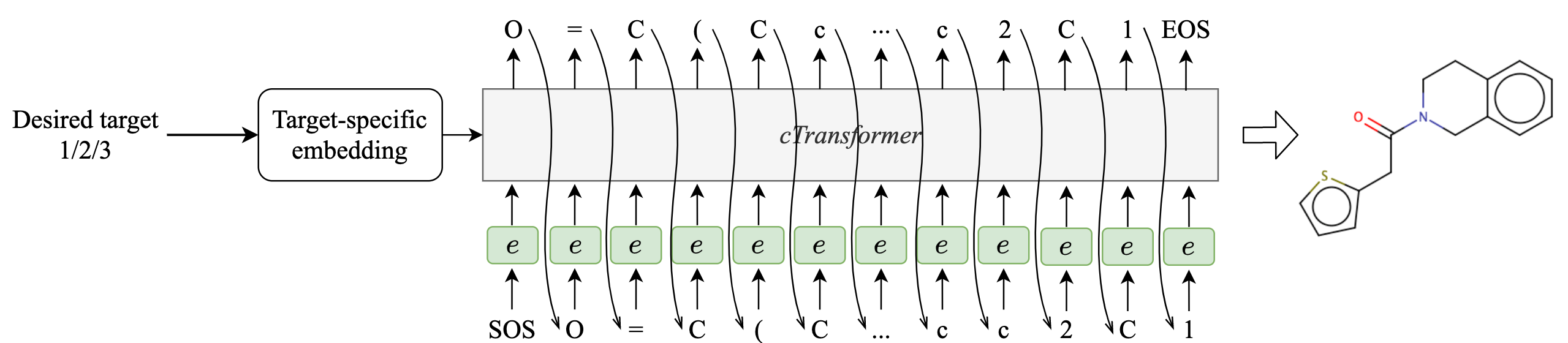}}
    \vspace{5pt}
    \caption{Target-specific conditional molecular generation.}
    \label{fig:generation}
\end{figure*}

\vspace{5pt}
\noindent\textbf{Incorporating target information} One of the major challenges in our task is to generate target-specific SMILE sequences. The model needs to not only memorize a valid drug-like structure, but also be able to memorize and generate target-specific information. The challenge lies in capturing target-specific information and further target-specific generation. We propose taking advantage of the Multi-head Attention in the Transformer decoder and imposing the target-specific embeddings to the \textit{keys} and \textit{values} of the attention operations. We denote our Transformer model with imposed conditional embeddings (e.g., target-specific embeddings) as \textit{cTransformer}.

We draw inspiration from a Transformer-based encoder-decoder structure for sequence-to-sequence translation. Taking machine translation as an example, the Transformer encoder-decoder model performs very well for this task. Specifically, the encoder memorizes the input sentence and stores them in ``memories'', and the decoder first attends to the previously generated tokens (\textbf{first mha}~\footnote{mha is multi-head attention}) and then performs multi-head attention over the output ``memories'' of the encoder (\textbf{second mha}). When attending to the ``memories'' of the encoder using \textbf{second} multi-head attention, the queries are from the \textbf{first} multi-head attention and \textit{values and keys} from the ``memories'' of the encoder. The intuition is to enable the decoder to attend over the input sequence~\cite{vaswani2017attention}.

We enable the SMILES sequence generation to condition on the specific target by feeding target-specific embeddings (denoted as $e^c$) to a decoder-only Transformer (shown in Figure~\ref{fig:training}-(b)). We use target-specific embeddings as \textit{keys and values} of \textbf{second mha}, which allows every position of the decoder to attend to the target-specific embeddings and ensures the subsequent token generations are conditioned on the target-specific embeddings. It is worth noticing that our target-specific design is orthogonal to the decoder and can be easily removed by setting the condition embeddings as zero embeddings. 

As shown in Figure~\ref{fig:training}, instead of fetching memory from an encoder, our decoder only design initialize the memory based on condition embeddings. For base model pre-training, $e^c$ is initialized with zero embeddings; when targets are involved, $e^c$ is initialized with target-specific embeddings.


\subsection{Workflow}
As in Figure~\ref{fig:training},
the training process of our task can be summarized as follows

\begin{enumerate}
    \item We first pre-train the \textbf{base model} of \textit{cTransformer} by setting the target-specific embeddings as zero embeddings (following Figure~\ref{fig:training}-(a) without feeding target-specific information) using data from Table~\ref{tab:data}-(a). We do not have any target constrain on the sequential generation and solely focus on learning the drug-like structure from the data.
    \item To feed target-specific information, we fine-tune \textit{cTransformer} with $<$compound, target$>$ pairs and enforce conditions of the corresponding target by feeding target-specific embeddings to the attention layer as ``memories'' (as shown in Figure~\ref{fig:training} (b)). We use data from Table~\ref{tab:data}-(b) where each SMILES sequence is manually tagged with a target (e.g., target proteins) indicating the specific physicochemical property of the small molecule. 
    \item We can generate drug-like structure by auto-regressively sampling tokens from the trained decoder (following Figure~\ref{fig:generation}). Optionally, we can enforce the desired target by feeding a target-specific embedding. The new generation will condition the target-specific information and likely has the desired property. 
\end{enumerate}



\subsection{Likelihood of molecular sequential generation}

An intuitive idea of likelihood estimation of SMILES string sampling is using Negative Log-Likelihood (NLL) loss. We propose a conditional design by enforcing \textit{values} and \textit{keys} of multi-head attention to be the generative condition (denoted as $\mathbf{c}$). Overall, our conditional NLL is as follows with the initial state set to $c$.


\begin{align*}
   NLL(S|c) = - & \bigg[ lnP(t_1|c) +  \sum^N_{i=2} ln P(t_i|t_{1:i-1},c) \bigg] 
\end{align*}

Where $c$ represents the generative condition (e.g., target protein), $S$ is a SMILES sequence with length $N$, and $t_i$ is the $i$-th token.  








\section{Related Work}


\subsection{Deep Learning-based Molecular Design}

Computational chemistry has reduced the experimental efforts of molecular design and overcome the experimental limitations~\cite{reymond2010chemical, cheng2012structure, scior2012recognizing, shoichet2004virtual}. With the emerging applications of deep learning, deep learning leads to a promising direction for drug discovery.

SMILES string is a popular representation of chemical structure in deep learning-based drug discovery. Recurrent Neural Network (RNN)-based generative models have been extensively tested in molecular design~\cite{yuan2017chemical, bjerrum2017molecular, gupta2018generative, segler2018generating}. \cite{arus2019exploring} samples an increased number of chemical structures by learning from a limited set of SMILES strings using Recurrent Neural Network (RNN). \cite{segler2018generating} further fine-tunes an RNN with a small set of SMILES strings, which are active against a specific biological target. Reinforcement learning can also further ~\cite{popova2018deep, olivecrona2017molecular} confine the chemical space for specific properties. 

Autoencoder is also applied in molecule generation~\cite{polykovskiy2020molecular,brown2019guacamol}.
A variational autoencoder can further generate molecules with specific properties by taking concatenated SMILES strings with the property of interests~\cite{lim2018molecular}. Our generation does not involve an encoder as we sample directly from the molecular structure memorized by the sequential decoder. 
RNNs can be further modified by setting the interval states of RNN cells (e.g., LSTM) to produce SMILES string with specific target properties~\cite{kotsias2020direct}. \cite{he2021molecular} proposes to modify an existing SMILES string based on the chemists' intention by prepending the desired changes to the original SMILES string. They test on both Transformer~\cite{vaswani2017attention} and RNN seq2seq structure by inputting a SMILES string along with the changed properties and outputting a SMILES string with the potentially desired properties.
In addition to variational autoencoder~\cite{lim2018molecular} and conditional recurrent neural networks~\cite{kotsias2020direct}, conditional graph generative models~\cite{li2018multi} also generate molecular graphs instead of SMILES with specified conditions, and the conditional representation is added as an additional term to the hidden state of each layer. 


\subsection{Linguistic Models}
Recurrent Neural Networks (RNNs) such as Gated Recurrent Unit (GRU)~\cite{GRU} and Long Short-Term Memory (LSTM)~\cite{LSTM} have successfully addressed a number of sequential modeling tasks and especially in language models. The idea of capturing temporal dependencies using RNN is by imposing connections to hidden units with a time delay to retain information about the past. Language models also benefit from RNN since we can treat a sentence of $m$ words as a sequence of $m$ time steps. In our sequential generation task, we need to memorize the previously generated tokens, also called context, to produce meaningful subsequent tokens. To be specific,  the prediction of the token at time step $t$ depends on the previous tokens generated during $t' < t$. RNN has an internal state for ``memorizing'' the context, which is updated by backpropagation every time step to reflect the context change. One of the drawbacks of recurrent structure is the difficulties in parallelization, which becomes more critical with longer input sequences. The idea of RNN also suffers from practical difficulties since the training of deep learning benefits from gradient descent, which becomes inefficient when the learning spans a long sequence of time steps~\cite{bengio1994learning}. In the task of generating SMILES sequences, the output sequence can expand to long sequences (e.g., the SMILES representation of Remdesivir has 92 characters).
RNN also suffers from the vanishing, and exploding gradient problems~\cite{bengio1994learning, pascanu2013difficulty} during training as the long-term gradients via back-propagation will tend to zero or infinity. Long short-term memory (LSTM)~\cite{LSTM} is proposed to address the  gradient problem but to some extend. 
In language models, the usage of RNN restricted the prediction to a short range while Transformer performs well in long-range predictions. Existing literature also experimentally ascertains that Transformer-based language models outperform LSTM in many tasks~\cite{GPT, DBLP:conf/iclr/LiuSPGSKS18}.

\vspace{5pt}
\noindent\textbf{Sequence-to-sequence generative model} The sequence-to-sequence~\cite{sutskever2014sequence} model (or seq2seq as it is commonly referred to) takes input in the form of a sequence of tokens and produces output also as a sequence. Such a model has been enjoying massive success in many natural language processing applications. The seq2seq modeling has been applied to many language tasks, such as text summarization~\cite{DBLP:conf/emnlp/RushCW15} and neural machine translation~\cite{DBLP:journals/corr/BahdanauCB14, wu2016google}, and the seq2seq models are trained end-to-end with a large number of training pairs (e.g., article-summary pairs for text summarization).  

Notable techniques in sequential modeling include attention mechanism~\cite{bahdanau2014neural} and pointer network~\cite{vinyals2015pointer}, which are designed for handling long sequences and rare tokens. Many successful applications can be seen in language modeling~\cite{merity2016pointer}, text summarization~\cite{gu2016incorporating}, text understanding~\cite{wen2017network}, and neural computing~\cite{graves2016hybrid}. Specifically, Transformer performs better than the standard LSTM encoder-decoder with attention in generating abstractive sections~\cite{DBLP:conf/iclr/LiuSPGSKS18}.

\begin{table*}[!h]
    \centering
    \begin{tabular}{l c c c c c c c c c c}
    \toprule
    \multirow{2}{*}{Model} & & & & \multicolumn{2}{c}{Frag$\uparrow$} & \multicolumn{2}{c}{SNN$\uparrow$} \\
    \cmidrule(r){5-8}
     & Valid$\uparrow$ &  Unique@1k$\uparrow$ & Unique@10k$\uparrow$ & Test & TestSF & Test & TestSF\\
    \midrule
    HMM & 0.076 & 0.623 & 0.567 & 0.575 & 0.568 & 0.388 & 0.38 \\
    NGram & 0.238 & 0.974 & 0.922 &0.985 & 0.982 & 0.521 & 0.5 \\
    Combinatorial & 1.0 & 0.998 & 0.991 &0.991 & 0.99 &0.451 & 0.439 \\
    CharRNN   & 0.975	& 1.0 & 0.999 & 1.0 &	0.998 & 0.601 & 0.565\\
    AAE &	0.937 &	1.0&	0.997 & 0.991	&0.99 &0.608	&0.568	\\
    VAE	& 0.977	&1.0	&0.998 &0.999	&0.998 & 0.626 &	0.578 \\
    JTN-VAE &	1.0	&1.0	& 1.0 & 0.997	&0.995&0.548	&0.519\\
    LatentGAN &	0.897	&1.0&	0.997 &	0.999 &	0.998&	0.538&	0.514 \\    \bottomrule
    \textbf{\textit{cTransformer}}$^*$ & 0.988 & \textbf{1.0} & 0.999 & \textbf{1.0} & \textbf{0.998}&0.619&	\textbf{0.578}\\
    \bottomrule
    \end{tabular}
    \vspace{5pt}
    \caption{Performance metrics for baseline models: fraction of valid molecules, fraction of unique molecules from 1,000 and 10,000 molecules. Fragment similarity (Frag) and similarity to a nearest neighbor (SNN) - results for random test set (Test) and scaffold split test set (TestSF). $^*$ means the base model of \textit{cTransformer}.}
    \label{tab:exp}
\end{table*}

\subsection{Generative Pre-trained Transformer}

A Generative Pre-trained Transformer (GPT)~\cite{GPT} is proposed for natural language processing (NLP) using Transformer and unsupervised learning. The intuitive idea is to train a model with a very large amount of data for natural language modeling in an unsupervised way and then fine-tune the pre-trained model with relatively small-sized supervised data for specific downstream tasks. The unsupervised pre-training is able to prime the model with drug-like knowledge and enforce valid SMILES strings.

GPT-2~\cite{GPT2}, a successor of GPT~\cite{GPT}, is a Transformer-based model that learns natural language by \textbf{predicting next word} given the previous words. GPT-2 is trained from a large text corpus using unsupervised learning. The well-trained GPT-2 is able to generate synthetic text excerpts \textbf{conditioning on an arbitrary input}, specifically the model generates a continuation of the input in the form of natural language. The synthetic text samples are conditioned on the input and adaptive to the style and content of the input. The benefits of the GPT-2 design are the capability of generating text (1) coherent with the input and (2) similar to human writing. GPT-2 has shown great capabilities of generating reasonable reviews by training on Amazon Reviews dataset conditioning on ratings and categories. 


\section{Experiment}

We evaluate the performance of base generative model in Section~\ref{sec:exp-base} and the performance of conditional generations in Section~\ref{sec:condition}.

\subsection{Base generative model}~\label{sec:exp-base}

\subsubsection{Dataset}
We use the MOSES molecular dataset from Polykovskiy \cite{polykovskiy2020molecular} to perform unsupervised pre-training on our \textit{cTransformer} with the target-specific embedding initialized as zero. It contains 1,760,739 drug molecules extracted from ZINC clean Lead Collection \cite{sterling2015zinc}, including 1,584,664 training molecules and 176,075 testing molecules. We follow the same train/test split as in \cite{polykovskiy2020molecular}.



\subsubsection{Evaluation}

In this section, we propose to use a set of metrics to quantitatively measure our generative model.
We evaluate the generated compounds in various aspects of molecule generation proposed in \cite{polykovskiy2020molecular}. Besides basic metrics such as chemical validity and diversity, we compare the distribution of drug-likeness properties between generated and real compounds. 

The performance is reported on 30,000 molecules generated from each generative model. We compute all metrics (except for validity) only for valid molecules from the generated set. We generate 30,000 molecules and use valid molecules from this set. We use the following metrics:

\begin{figure*}
    \centering
    \resizebox{!}{.6\textwidth}{
    \includegraphics{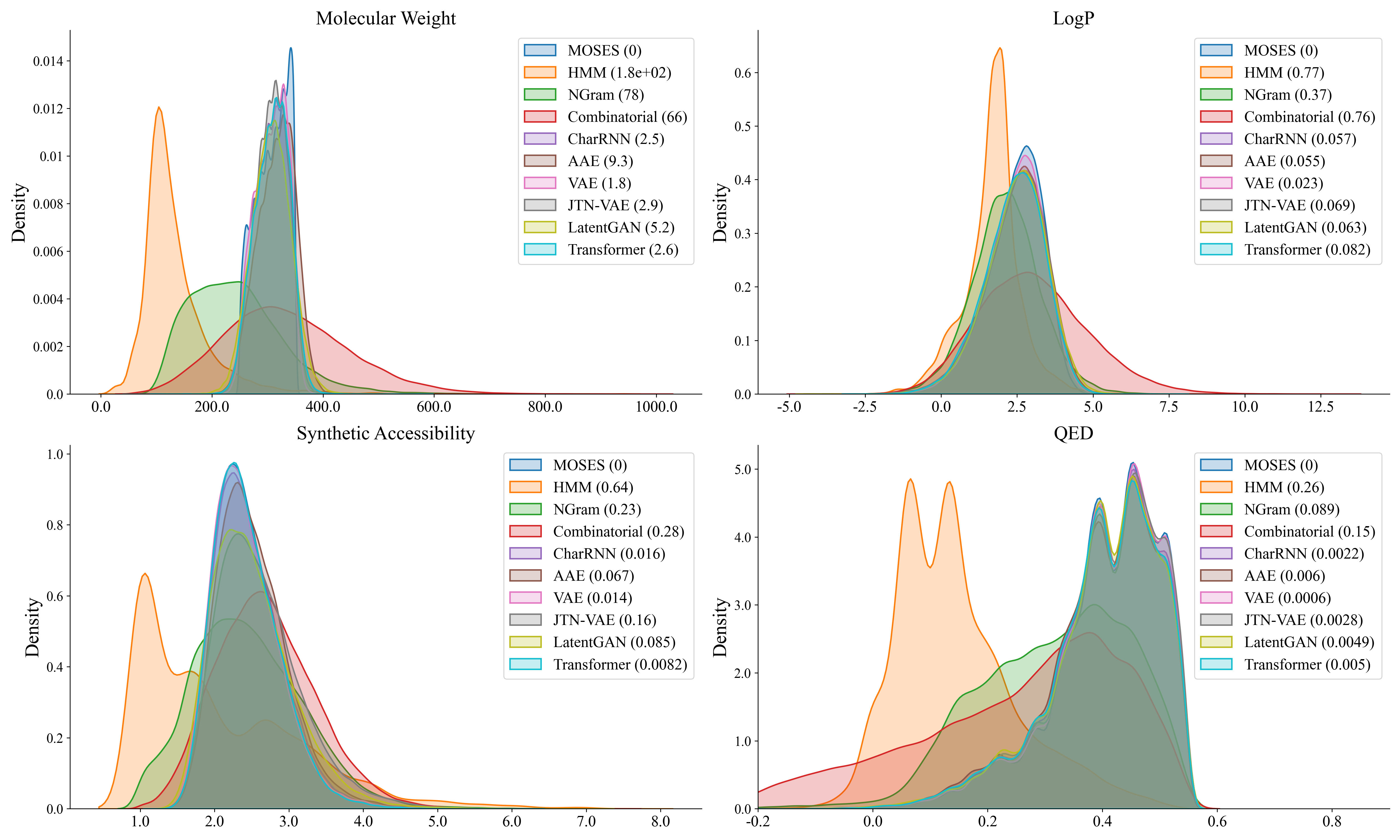}}
    \caption{Distribution of chemical properties for MOSES dataset and sets of generated molecules. Wasserstein-1 distance to MOSES test set is denoted in parenthesis. We cover Molecular Weight, LogP, Synthetic Accessibility, and QED.}
    \label{fig:vis_properties}
\end{figure*}

\begin{itemize}
\item \textbf{Fraction of valid (Valid) and unique molecules} report valid and unique SMILES strings. The validity is checked using a molecular structure parser (RDKit).

    \item \textbf{Unique@1K} and \textbf{Unique@10K} for the first $1000$ and $10000$ valid molecules in the generated set.  In general, validity measures whether the model captures enough chemical constraints (e.g., valence); Uniqueness measures whether the model overlaps with trained molecules.

    \item \textbf{Fragment similarity (Frag)} compares distributions of BRICS fragments~\cite{degen2008art} in generated and reference sets. This metric measures how similar are the scaffolds present in generated and reference datasets. 

    \item \textbf{Similarity to a nearest neighbor (SNN)} calculates the average Tanimoto similarity (also known as the Jaccard index) between fingerprints of a molecule from the generated set and its nearest molecule in the reference dataset based on ~\cite{rogers2010extended}~\cite{rdkit}. 

\end{itemize}

We compare our method with the following \textbf{baselines}:
\begin{itemize}
    \item Hidden Markov Model (HMM) uses Baum–Welch algorithm to learn a probabilistic distribution over the SMILES strings, then samples the next token and state from learned probabilities. 
    \item N-gram generative model (NGram)~\cite{polykovskiy2020molecular} collects statistics of n-grams frequencies in the training set and uses such a distribution to  sample strings. 
    \item Combinatorial generator (Combinatorial)~\cite{polykovskiy2020molecular} splits molecules into fragments and generates new molecules by random connections. The sampling of fragments is based on training data frequencies.
    \item Character-level recurrent neural network (CharRNN)~\cite{segler2018generating} learns the probability distribution of the next token over the vocabulary space conditioning on previously generated tokens using RNN.
    \item Variational autoencoder (VAE)~\cite{kingma2013auto} encodes SMILES strings to the latent space, then reconstructs the strings from the latent codes with a decoder. New SMILES strings can be sampled from the latent space.
    \item Adversarial Autoencoder (AAE)~\cite{makhzani2015adversarial} replaces the objective of VAE with an adversarial objective, which enables relaxed assumptions of the prior distribution.
    \item Junction Tree VAE (JTN-VAE)~\cite{jin2018junction} generates molecules by first generating a tree-structured object and further assembled into a molecule.
    \item Latent Vector Based Generative Adversarial Network (LatentGAN)~\cite{prykhodko2019novo} pre-trains an autoencoder to map SMILES into the latent space and mapped back to SMILES, then uses a generative adversarial network to produce the latent code.
\end{itemize}

The performance of the different approaches is summarized in Table~\ref{tab:exp}. Our method (base model of \textit{cTransformer}) achieves state-of-the-art results in the Fraction of valid (Valid), Unique@1k, Unique@10k, Fragment similarity, and Similarity to the nearest neighbor. 

We also compare the distribution of chemical properties for MOSES dataset and sets of generated molecules using the following chemical properties: 
\begin{itemize}
     \item \textbf{Molecular weight (MW)} is the sum of atomic weights in a molecule. 
    \item \textbf{LogP} is the octanol-water partition coefficient computed by RDKit’s Crippen~\cite{wildman1999prediction} estimation.
     \item \textbf{Synthetic Accessibility Score (SA)} estimates the ease of synthesis (synthetic accessibility) of drug-like molecules based on molecular complexity and fragments contributions~\cite{ertl2009estimation}. 
     \item \textbf{Quantitative Estimation of Drug-likeness (QED)} measures the drug-likeness based on desirability in a value between zero (all properties unfavorable) to one (all properties favorable)~\cite{bickerton2012quantifying}.  
\end{itemize}

The resultant distributions of four molecular properties in generated and test datasets are shown in Figure~\ref{fig:vis_properties}. Our model closely matches the real data distribution. This validates that our method is capable of generating drug-like molecules.

\begin{figure*}
    \centering
    \resizebox{!}{0.8\textwidth}{
    \includegraphics{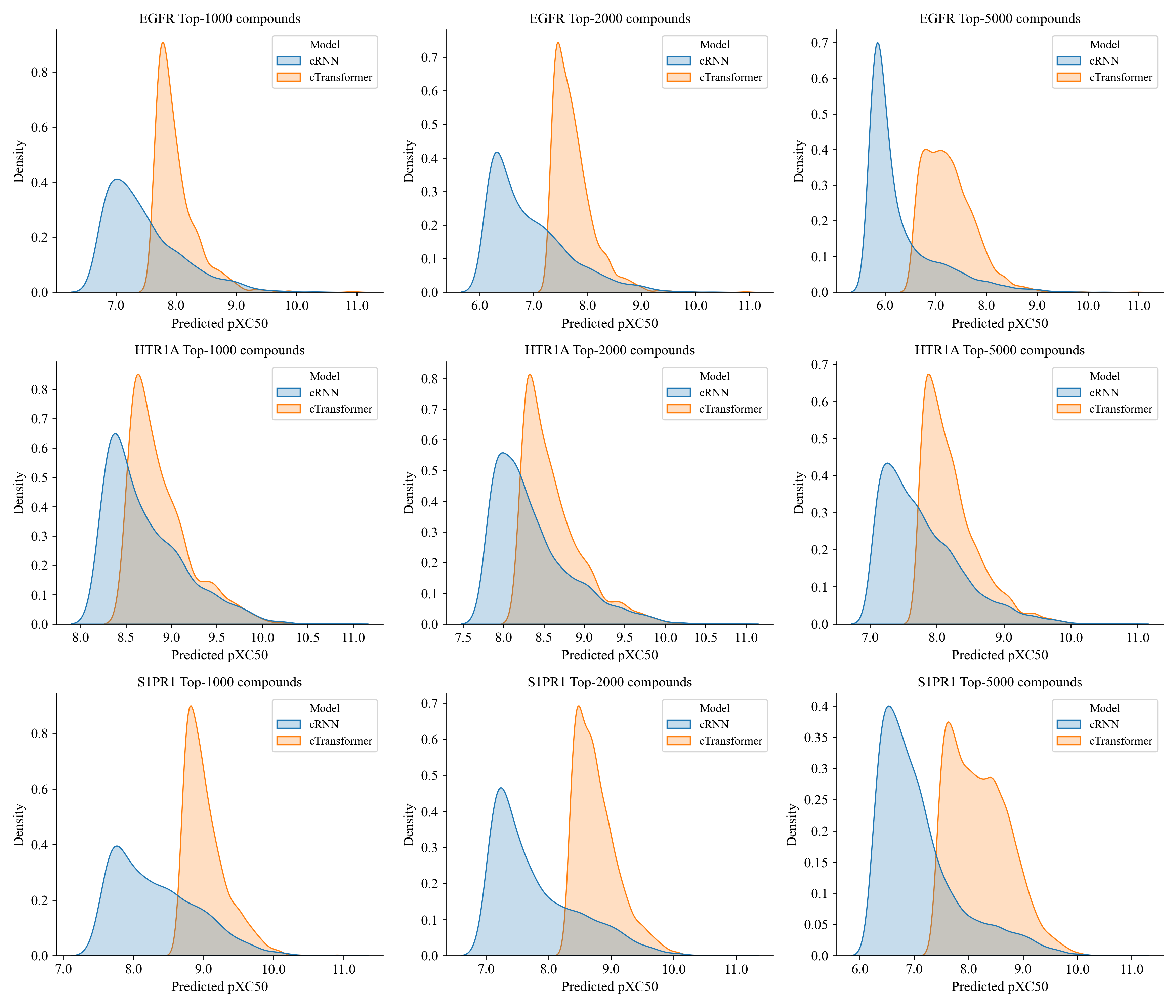}}
    \caption{Distribution of predicted activity (pXC50) of the top 1000 (left) / 2000(middle) / 5000(right) compounds from cTransformer(orange) and cRNN(blue) for EGFR(top), HTR1A(middle) and S1PR1 (bottom) targets.}
    \label{fig:dis}
\end{figure*}

\subsection{Target-specific molecular generation}~\label{sec:exp-condition}

\begin{figure*}
    \centering
    \resizebox{!}{0.6\textwidth}{
    \includegraphics{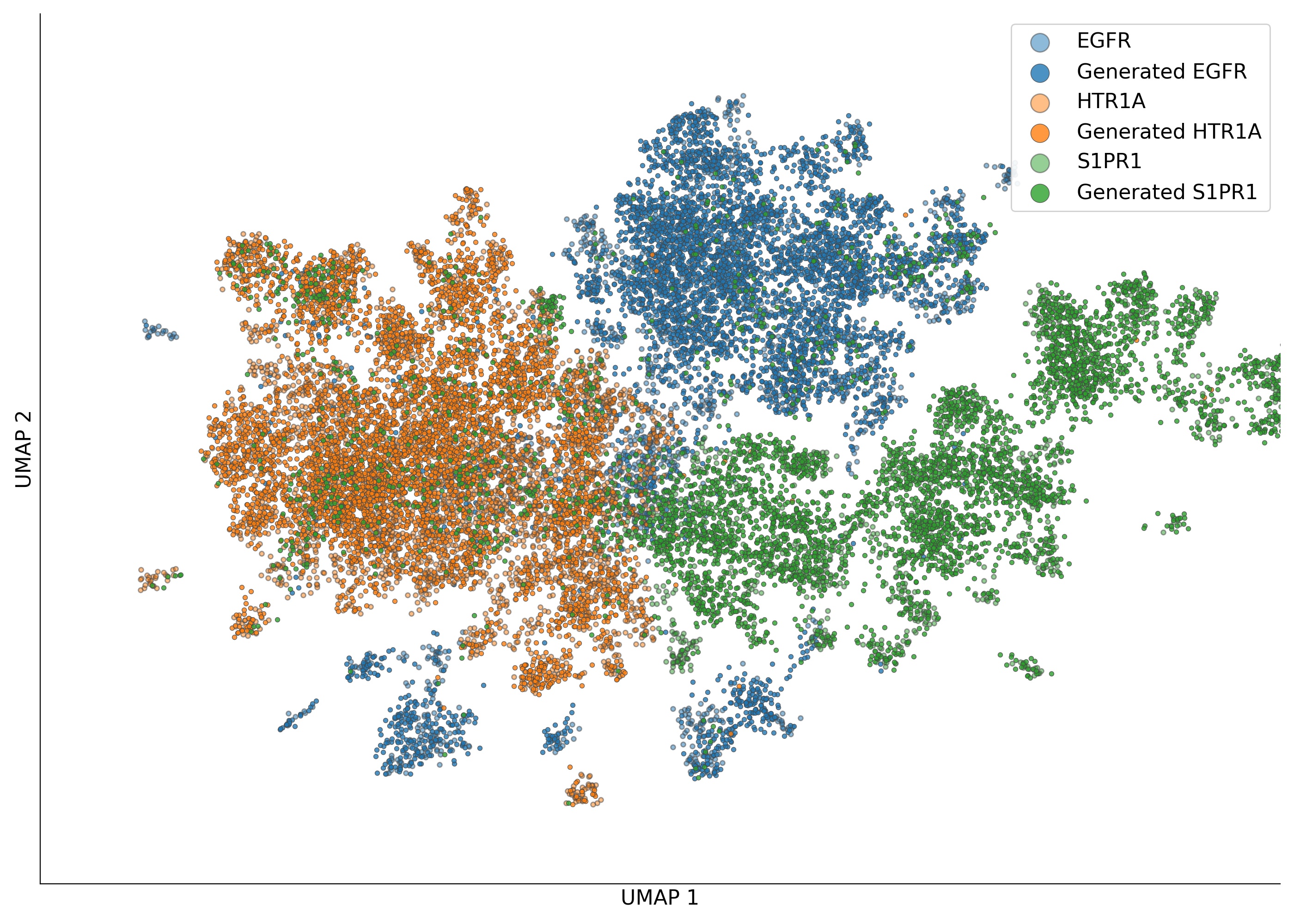}}
    \vspace{5pt}
    \caption{UMAP of the generated target-specific molecules (dark color) by the \textit{cTransformer} and the ground-true target-specific molecules (light color).}
    \label{fig:vis-compare}
\end{figure*}

\subsubsection{Dataset}

\begin{table}[!b]
    \centering
    \begin{tabular}{l c c c c c }
    \toprule
     \multirow{2}{*}{Target} & & & & \multicolumn{2}{c}{QSAR} \\
    \cmidrule(r){5-6}
     & \# of active mols &  \# of mols & Activity & R & RMSE \\
    \midrule

    EGFR & 1,381 &	5,181	&	$6.29 \pm 1.39$ & $0.843$ & $0.588$ \\
    HTR1A & 3,485	&	6,332	&	$7.33	\pm	1.23$ & $0.763$ & $0.631$    \\
    S1PR1 &	 795 &	1,400	&	$7.25	\pm	1.58$ & $0.825$ & $0.779$ \\
    \bottomrule
    \end{tabular}
    \caption{Target data sets and the performance of the QSAR models. The active compounds are used for training target-specific generative models. QSAR models are trained on both active and non-active compounds for each target. R stands for Pearson correlation. RMSE stands for root mean square error.}
    \label{tab:target-dataset}
\end{table}

We download the target-specific molecular dataset from \cite{prykhodko2019novo} for training our target-specific \textit{cTransformer}, which contains $1,381$, $795$, and $3,485$ active molecules corresponding to EGFR, S1PR1, and HTR1A target protein, respectively. 

\subsubsection{ML-based QSAR model for active scoring}
To evaluate the capability of generating active target-specific compounds, we build a regression-based QSAR (Quantitative structure-activity relationship) model for each target. The molecular set with activity data for each target is acquired from ExCAPE-DB\cite{sun2017excape}, which includes $5,181$, $6,332$, and $1,400$ molecules corresponding to EGFR,  HTR1A, and S1PR1 target proteins, respectively. We train a LightGBM model to predict activity on $2,533$ molecular features, including a 2048-length FCFP6 fingerprint, 166-length MACCSkeys, and 319 RDKit Molecular Descriptors. We report Pearson correlation and RMSE on test dataset in table \ref{tab:target-dataset}. All the testing Pearson correlations of QSAR models are higher than $0.75$, demonstrating that our QSAR models are capable of benchmarking the activity of the generated compounds.

\begin{table}[!t]
    \centering
    \begin{tabular}{l c c c c  }
    \toprule
                Target  & Model   & Valid$\uparrow$ &  Unique@10k$\uparrow$ & Novel $\uparrow$ \\
    \midrule
\multirow{2}{*}{EGFR} &{cRNN} & 0.921 &  0.861 & 0.662   \\
                  &{cTransformer}  & 0.885 & \textbf{0.940} & \textbf{0.898}   \\
\multirow{2}{*}{HTR1A} &{cRNN}  & 0.922 & 0.844  &  0.498  \\
                  &{cTransformer} &	0.905 & \textbf{0.896} &   \textbf{0.787}  \\
\multirow{2}{*}{S1PR1}  &{cRNN} & 0.926 & 0.861   & 0.514 \\
                  &{cTransformer} & 0.926 & 0.838 &   \textbf{0.684}   \\
    \bottomrule
    \end{tabular}
    \caption{Evaluation metrics: fraction of valid molecules, fraction of unique molecules from $10,000$ molecules, and novelty (fraction of molecules not present in the training set)}
    \label{tab:target-res}
\end{table}

\subsubsection{Evaluation}
To benchmark the performance of the targeted models, we built a conditional RNN model (denoted as \textit{cRNN}) by first training a base RNN model on the same MOSES set and fine-tuning it on the target set. 

\begin{figure*}
    \centering
     \resizebox{!}{1.3\textwidth}{
    \includegraphics{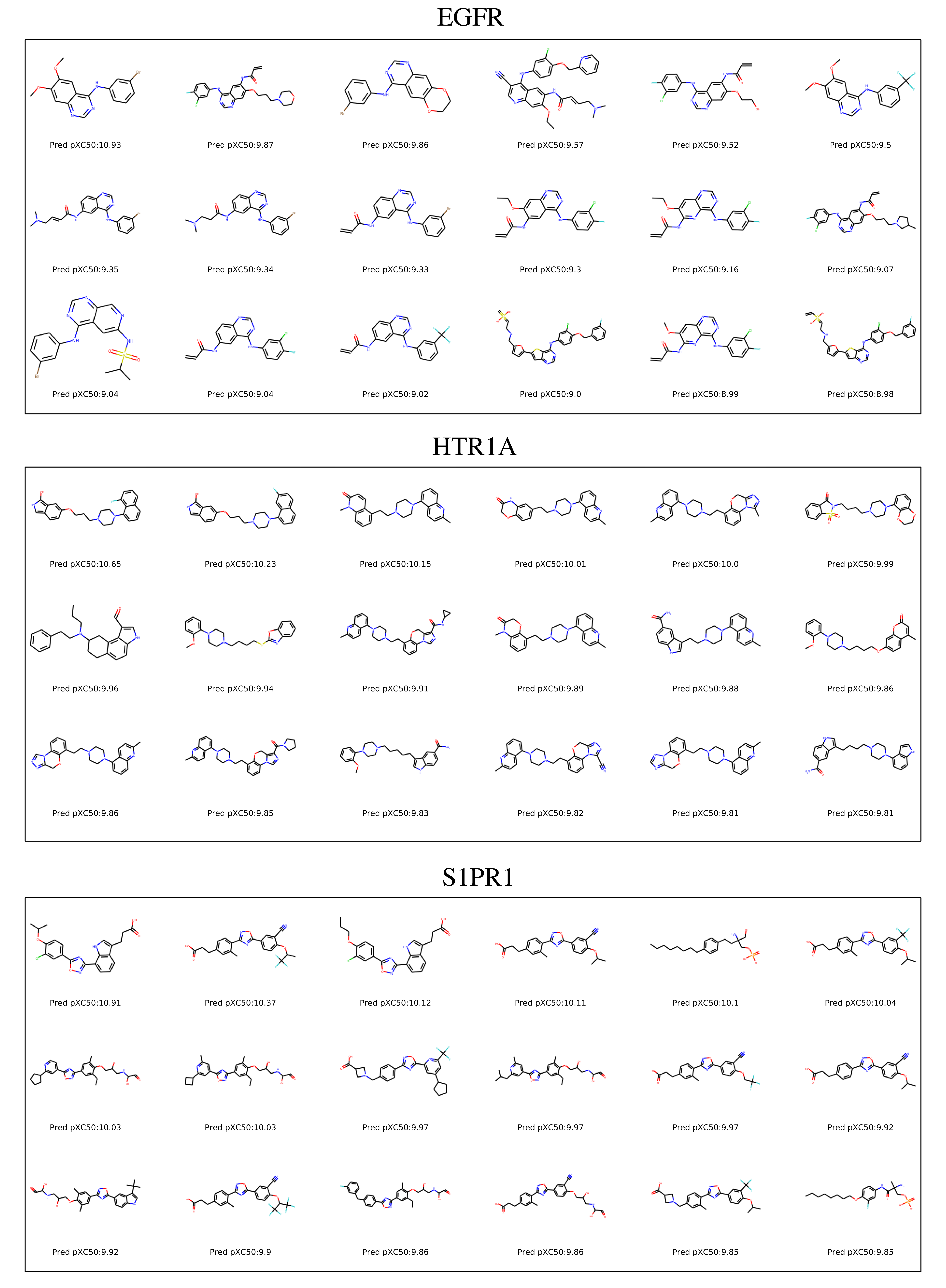}}
    \caption{The top 24 most active generated compounds for each target protein.}
    \label{fig:examples}
\end{figure*}

We sampled 30,000 compounds from the \textit{cTransformer} and \textit{cRNN} model and reported the metric in the table \ref{tab:target-res}. Results demonstrate that the validity of all the cases was above $88\%$, and the uniqueness of 10k valid compounds was $94\%$, $90\%$, and $83\%$ for EGFR, HTR1A, and S1PR1, respectively, and are higher than the values of cRNN except for S1PR1. Moreover, in terms of novelty, the values are $90\%$, $78\%$, and $68\%$ for EGFR, HTR1A, and S1PR1 respectively, which significantly outperforms the \textit{cRNN} model. This shows that \textit{cTransformer} is not only able to generate valid compounds but also design novel molecules, which is critical for de novo drug design. Furthermore, we use QSAR models to predict the activity of all the generated valid compounds. We rank the compounds based on predicted value and plot the activity distribution of the top $1000$/$2000$/$5000$ most active ones in Figure~\ref{fig:dis}. Results show that the distributions of predicted activity of compounds from \textit{cTransformer} are significantly better than those from \textit{cRNN} for all three targets. This highlights that \textit{cTransformer} is more capable of generating active compounds than \textit{cRNN}. Additionally, we show the top 24 most active molecules in Figure~\ref{fig:examples}. 

Moreover we evaluate the capability of generating target-specific compounds by visualizing the chemical space. The hypothesis is that the compounds that can potentially interact with the same protein target would populate the same sub-chemical space. To evaluate the overlapping of chemical space, we select top predicted active 5000 compounds for each target, then the 1024-bit FCFP6 fingerprint vectors are calculated for the generated compounds and the real compounds from the training dataset (the training dataset here refers to the fine-tuning dataset in Table~\ref{tab:data}-(b)). We use the Uniform Manifold Approximation and Projection (UMAP) to construct 2D projections. These projections are illustrated in Figure~\ref{fig:vis-compare}. Each point corresponds to a molecule and is colored according to its target label. The \textit{dark} and \textit{light} colors represent the \textit{generated} compounds and \textit{training set} compounds, respectively. The visualization of chemical space in Figure~\ref{fig:vis-compare} demonstrates the generated target-specific molecules (dark color) and real target-specific molecules (light color) occupy the same sub-chemical space. These results show that our \textit{cTransformer} can generate compounds that are similar to the ones in the training set but are still novel structures.

\section{Conclusion and Future Work}

In this study, we first present a Transformer-based random molecular generator and compare it with a number of baseline models using standard metrics. We demonstrate that the Transformer-based molecular generation achieves state-of-the-art performances in generating drug-like structures. To incorporate the protein information, we present a target-specific molecular generator by feeding the target-specific embeddings to a Transformer decoder. We apply the method on three target-biased datasets (EGFR, HTR1A, and S1PR1) to evaluate the capability of the cTransformer to generate target-specific compounds and compare it with conditional RNN. Our results demonstrate that the sampled compounds from the model are predicted to be more active than \textit{cRNN} in all three targets. Additionally, we visualize the chemical space, and the generated novel target-specific compounds largely populate the original sub-chemical space. In summary, these experimental results demonstrate that our \textit{cTransformer} can be a valuable tool for de novo drug design.

\balance


\bibliography{ref}
\bibliographystyle{unsrt}

\end{document}